\documentclass[sigconf]{acmart}

\AtBeginDocument{%
  \providecommand\BibTeX{{%
    \normalfont B\kern-0.5em{\scshape i\kern-0.25em b}\kern-0.8em\TeX}}}

\setcopyright{none}
\renewcommand\footnotetextcopyrightpermission[1]{}
\thanks{This work was supported by the National Science Foundation Award IIS-2220956 and the University of Minnesota Informatics Institute (UMII) MnDRIVE Graduate Assistantship Award.}
\thanks{The authors are with the department of Computer Science \& Engineering and the Minnesota Robotics Institute, University of Minnesota - Twin Cities, MN, USA. Email: {\tt\small \{$^{1}$enan0001, $^{2}$junaed\}@umn.edu}}

\newcommand{\eg}{\emph{e.g.}, }
\newcommand{\ie}{\emph{i.e.}, } 
 
\usepackage[inline]{enumitem}
\usepackage{subcaption}
\usepackage{graphicx}
\usepackage{multirow}
\usepackage{caption}
\usepackage{hyperref}
\usepackage{diagbox}
\usepackage{textcomp}

\begin{document}

\title{A Diver Attention Estimation Framework for Effective Underwater Human-Robot Interaction}

\author{Sadman Sakib Enan$^1$ and Junaed Sattar$^{2}$}

\begin{abstract}
Many underwater tasks, such as cable-and-wreckage inspection and search-and-rescue, can benefit from robust Human-Robot Interaction (HRI) capabilities. With the recent advancements in vision-based underwater HRI methods, Autonomous Underwater Vehicles (AUVs) have the capability to interact with their human partners without requiring assistance from a topside operator. 
However, in these methods, the AUV assumes that the diver is ready for interaction, while in reality, the diver may be distracted.
In this paper, we attempt to address this problem by presenting a diver attention estimation framework for AUVs to autonomously determine the \textit{attentiveness} of a diver, and developing a robot controller to allow the AUV to navigate and reorient itself with respect to the diver before initiating interaction.
The core element of the framework is a deep convolutional neural network called DATT-Net. It is based on a pyramid structure that can exploit the geometric relations among 10 facial keypoints of a diver to estimate their head orientation, which we use as an indicator of attentiveness.
Our on-the-bench experimental evaluations and real-world experiments during both closed- and open-water robot trials confirm the efficacy of the proposed framework.
\end{abstract}

\maketitle
\thispagestyle{empty}
\pagestyle{empty}

\section{INTRODUCTION}
In recent years, AUVs have been used in independent missions (\eg underwater oil spill survey system~\cite{vasilijevic2015heterogeneous}, oceanographic survey system~\cite{hwang2019auv}), and also increasingly in collaborative settings containing both humans and other AUVs (\eg \cite{xia2019visual, islam2018understanding}).
The ever-increasing on-board computing power, availability of low-cost and affordable AUVs, and improved HRI capabilities are some of the contributing factors behind the surge in AUV usage. Currently, humans and AUVs can effectively interact with each other using a number of different methods, such as high-speed tethered communication~\cite{aoki1997development}, fiducial markers~\cite{sattar2007fourier}, gesture-based framework~\cite{islam2018dynamic}, small displays and lighting schemes~\cite{demarco2014underwater}, and motion-based methods~\cite{fulton2022robot}. 
But these interaction systems usually require both the diver and the AUV to be oriented properly and to be in close proximity to each other. And, almost all of the time it is the diver who swims towards the AUV, orients themselves correctly with respect to the AUV, and begins the interaction. However, this puts additional physical and cognitive burdens on the divers. On the other hand, AUVs do not currently have this capability as it is extremely difficult for a machine to intelligently perceive human orientation and attention underwater, localize itself and the target, navigate towards the target, and reorient itself all at the same time. This is mainly due to the domain-specific challenges underwater, \eg degraded visuals, reduced sensing capability arising from signal attenuation, less reliable localization and motion planning, and lack of high-bandwidth wireless communication. Nevertheless, AUVs are successfully using their visual perception capabilities in littoral regions owing to the success of underwater vision improvements~\cite{fabbri2018enhancing} and novel deep learning innovations~\cite{islam2020sesr}. This success contributes to the development of the work in~\cite{fulton2022using}, which proposes to use monocular cameras to approach a human diver by leveraging the pose of the body and biological priors to facilitate human-robot interaction. This method does not require the use of expensive sensors, such as stereo cameras, sonar, acoustic sensors (\eg Doppler Velocity Logs), and localization devices, and yet achieves promising results. However, the proposed solution always assumes that a diver is ready to interact. There are times when a diver may be engaged in a different task or interaction with a fellow dive buddy but is still willing to interact with a robot only if the robot positions itself conveniently. In such scenarios, it is imperative for the AUVs to understand the intent of the diver, more specifically their attentiveness. \textit{In the context of the paper, we use the phrases `diver attention' and `attentiveness of the diver' interchangeably, which refer to whether or not a diver is looking at a robot with an intention to interact and/or collaborate.}
\begin{figure}[t]
    \centering
    \vspace{2mm}
    \includegraphics[width=.99\linewidth]{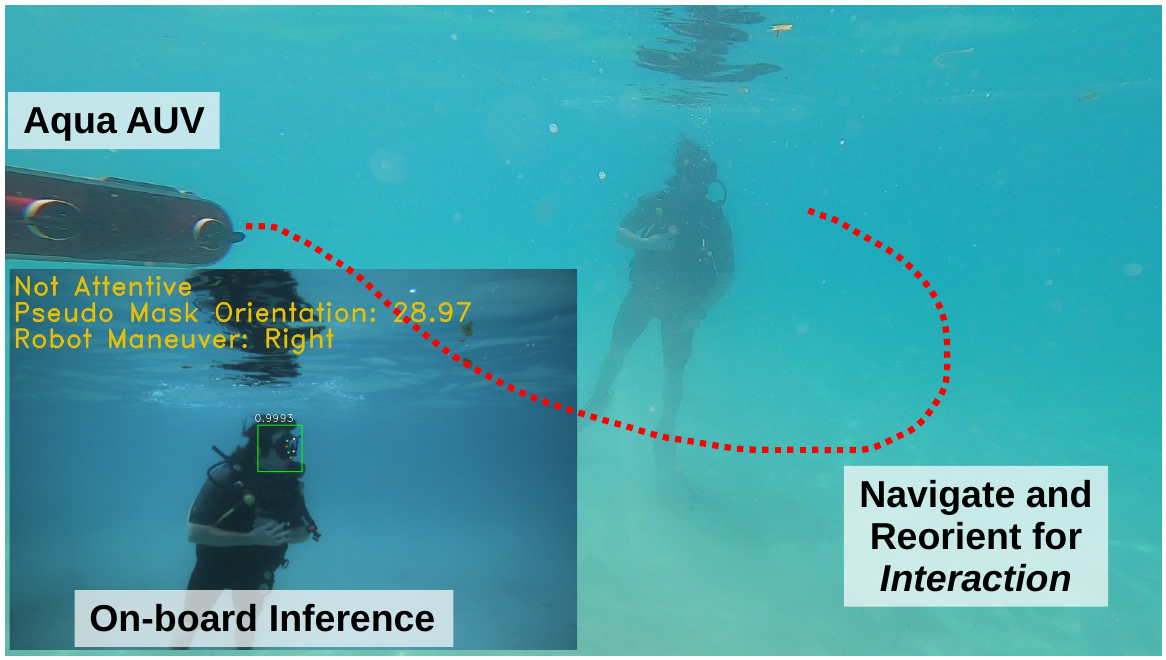}
    \caption{Demonstration of the proposed framework running on-board the Aqua AUV~\cite{dudek2007aqua} during our open-water experiments in the Caribbean Sea, off the coast of Barbados. The red dashed line represents the robot's future trajectory to position itself conveniently for interaction.}
    \label{fig:intro}
    \vspace{-4mm}
\end{figure}

In this work, we present a novel diver attention estimation framework which will enable an AUV to detect the attentiveness of a diver. If the diver is attentive, the AUV signals its intention to begin interaction immediately. If the diver is not
attentive, the AUV navigates and reorients itself as necessary to show its intention to interact with the
diver. To determine the attentiveness of the diver, however, the AUV needs to know the head orientation or the head pose of the diver. In terrestrial applications, researchers solve 3D (head) pose estimation problems using various learning algorithms (\eg\cite{ruiz2018fine, liu20163d, meyer2015robust}), which use datasets that contain 3D locations of the pose keypoints with respect to a global frame. However, there are no datasets that contain 3D head pose data of scuba divers, partly arising from the difficulty and logistical challenges in collecting such datasets. In contrast, some researchers model the 3D pose estimation problem as a 2D problem, and have used keypoints regression to address the 3D pose estimation challenge~\cite{andriluka2010monocular, chen20173d}. However, it is almost impossible to estimate the 3D pose from a 2D image if the corresponding 2D keypoints are not even visible in the image space. Consequently, the unseen keypoints are not even annotated in the datasets and are not used in establishing the 2D-3D correspondence. In our work, we propose to tackle this unique problem by creating the very first \underline{D}iver \underline{Att}ention (\textbf{DATT}) dataset which contains a wide collection of annotated diver images. We carefully annotate the diver face bounding box and $10$ facial keypoints in a way that the keypoints represent the orientation of the head in the image space. In fact, we annotate the facial keypoints even if the frontal face is not fully visible to promote learning of the geometric structure of the keypoints, as opposed to only the individual keypoints location. In addition, we design a deep neural network architecture (\textbf{DATT-Net}) which can be supervised to learn the spatial location of the $10$ facial keypoints. Finally, we use the detected keypoints to estimate the attentiveness of a scuba diver and design a robot controller for the Aqua AUV~\cite{dudek2007aqua} to navigate and reorient itself with respect to a diver to begin an interaction (as shown in Fig.~\ref{fig:intro}). 

We make the following contributions in this paper:
\begin{enumerate}
    \item We present the DATT dataset containing $3,314$ annotated diver images, collected during multiple closed-water robot trials. The images are carefully annotated to include diver face bounding boxes and $10$ facial keypoints of divers. 
    \item We propose an end-to-end deep neural network called DATT-Net to capture diver faces and their associated facial keypoints even if a diver is looking at a $90$ degree angle with respect to the camera. The training pipeline includes a multi-loss objective function which gives emphasis on maintaining the actual geometric relation among the facial keypoints. 
    \item We perform spread analysis on the facial keypoints to determine the attentiveness of a diver and define a $pseudo$-angle variable to quantify the direction of diver's attention.
    \item We perform a number of qualitative and quantitative experiments that validate that the proposed framework can robustly estimate the attentiveness of the diver. 
    \item We deploy the proposed algorithm on a physical AUV and further design a controller which helps navigate and reorient the AUV to position itself for interaction with a human diver when required.
\end{enumerate}
\section{Related Work}
AUVs are successfully using their visual perception capabilities in both littoral and offshore regions owing to the success of underwater vision improvements (\eg\cite{fabbri2018enhancing,islam2020underwater}) and novel deep learning innovations (\eg\cite{islam2020sesr,islam2022svam}). This success contributes to the development of the work in~\cite{fulton2022using}, which proposes to use monocular cameras to approach a scuba diver by leveraging their body pose and biological priors to facilitate HRI. This method does not require the use of expensive sensors, such as stereo cameras, sonar, acoustic sensors (\eg Doppler Velocity Logs), and localization devices, and yet achieves promising results. However, the proposed solution always assumes that a diver is ready to interact. There are times when a diver may be engaged in a different task or interaction with a fellow dive buddy but is still willing to interact with a robot only if the robot positions itself conveniently. In such scenarios, it is imperative for the AUVs to understand the intent of the diver, more specifically their attentiveness.

To determine the attentiveness of the diver, the AUV needs to know the head orientation or head pose of the diver. In terrestrial applications, researchers solve 3D head pose estimation problems using various learning algorithms (\eg\cite{ruiz2018fine, liu20163d, meyer2015robust}), which use datasets that contain 3D locations of the pose keypoints with respect to a global frame. However, there are no datasets that contain 3D head pose data of scuba divers, partly arising from the difficulty and logistical challenges in collecting such datasets. In contrast, some researchers model the 3D pose estimation problem as a 2D problem and have used keypoints regression to address the 3D pose estimation challenge (\eg\cite{andriluka2010monocular, chen20173d}). But it is almost impossible to estimate the 3D pose from a 2D image if the corresponding 2D keypoints are occluded in the image space. Consequently, the unseen keypoints are not even annotated in the datasets and are not used in establishing the 2D-3D correspondence. The works in~\cite{gomez2019caddy} and~\cite{huang2022human} tackle the diver pose estimation problem to estimate diver heading (\ie orientation), however, they either require stereo image pair or use a computationally expensive model that is not suitable for real-world deployments. 
We propose to tackle these issues by creating the DATT dataset where we carefully annotate images captured by monocular cameras with diver face bounding boxes and $10$ facial keypoints. We annotate the facial keypoints even when the frontal face is occluded, to promote learning of keypoints' geometric structure rather than just their individual locations.
\section{DATT Dataset}
To facilitate the training and testing of DATT-Net, we prepare a dataset that contains a wide variety of annotated underwater diver images. The images have resolutions of $1920\times1080$ pixels and are collected from multiple closed-water environments using GoPros~\cite{gopro}. 
There are a total of $3,314$ annotated diver images (training set: $2,652$ and test set: $662$). To increase the diversity of the dataset, the divers were instructed to appear differently underwater. Some divers wore scuba masks while others used pairs of goggles. Some used their snorkels while a few others did not. The head orientation of the divers were also varied by instructing them to look directly at the camera or to look elsewhere at different angles (both in the left and right directions). As a result, there are even instances where the diver faces are occluded from the camera's perspective, such as when they looked $90$-degrees to the left or right.
\begin{figure}[t]
    \centering
    \vspace{2mm}
    \includegraphics[width=.95\linewidth]{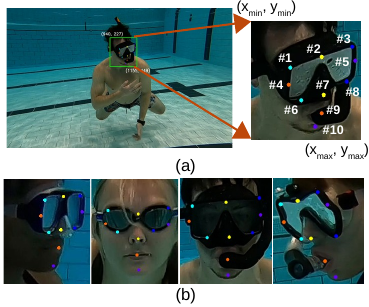}
    \caption{(a) A sample image from the DATT dataset and the corresponding annotations. Each annotation consists of a bounding box label, $\Vec{b}=[x_{min},y_{min},x_{max},y_{max}]$, for the face, and a label for $10$ facial keypoints denoted by, $\Vec{p}=[x_1, y_1, \cdots, x_{10}, y_{10}]$. (b) A few additional annotated samples where the divers are wearing different types of scuba gear and are looking at different directions.}
    \label{fig:data}
    \vspace{-4mm}
\end{figure}
\begin{figure*}[t]
    \centering
    \vspace{2mm}
    \includegraphics[width=.95\linewidth]{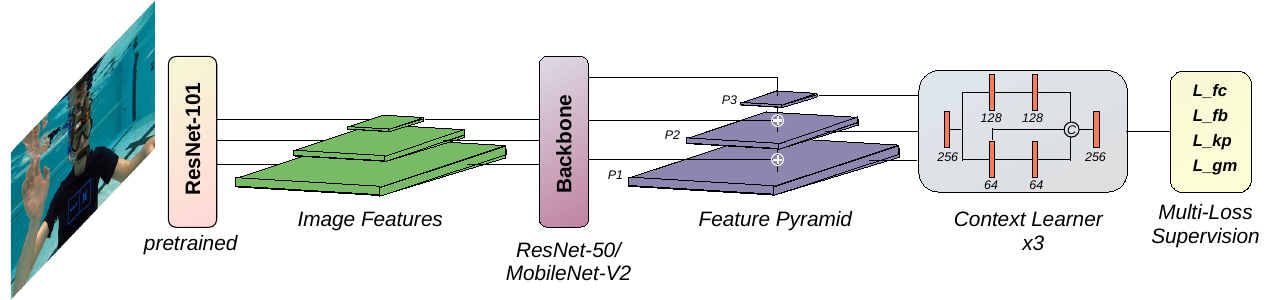}
    \caption{The network architecture of DATT-Net employs multi-scale learning on a feature pyramid and includes an additional supervision branch to learn the geometric relations of $10$ facial keypoints. This allows the network to operate effectively even when the diver is facing the robot at a $90$-degree angle.}
    \label{fig:model}
\end{figure*}

Since we want to estimate the head orientation of the divers directly from the 2D images, we carefully annotate the images to reflect as many 3D perspective properties as possible. We start by annotating the diver face bounding boxes. To estimate the head orientation of the diver, a full set of facial keypoints and their underlying geometric relations need to be used. Traditionally, researchers select eyes and two corners of the lips, among others, as three of the vital facial keypoints for terrestrial applications. However, for a diver underwater, their eyes are usually obscured by the scuba mask or goggles and the shape of their lips are mostly distorted by the snorkels or regulators. So, instead of selecting facial keypoints in the traditional manner, we leverage the following facts. Since the scuba gear on the diver's face (\eg masks and regulators) are rigid objects, we select the majority of our facial keypoints on them and focus mainly on finding the geometry of the keypoints that preserves the shapes of the scuba gear in 3D space. Specifically, we select $10$ facial keypoints which are easier to identify underwater compared to the traditional facial keypoints, where eight keypoints are on the mask, one is on the snorkel or regulator, and the remaining keypoint is on the diver's chin. We always annotate all $10$ facial keypoints even when some of them are occluded from the camera’s perspective so that a downstream algorithm can regress the keypoints based on the actual geometric relations among the keypoints. Fig.~\ref{fig:data} shows a few annotations from DATT dataset. We make it available at \url{https://irvlab.cs.umn.edu/datt}.

\section{DATT-NET}
The primary goal of the proposed diver attention estimation framework is to estimate the head orientation of a scuba diver to determine if they are attentive to a companion robot. To achieve this, we need to- \begin{enumerate*}
    \item localize the diver's face,
    \item detect the facial keypoints of the diver, and
    \item determine the attentiveness of the diver from the geometry of the keypoints.
\end{enumerate*}
We propose to solve the first two tasks with a deep convolutional neural network called DATT-Net, which employs a pyramid structure in its design, similar to the ones shown in FPN~\cite{lin2017feature} and RetinaFace~\cite{deng2020retinaface}. The key components of DATT-Net are described below.

\subsection{Feature Extractor}
DATT-Net uses a ResNet-101 block~\cite{he2016deep}, pretrained on ImageNet dataset~\cite{deng2009imagenet}, to extract the initial features from the input image of $640\times640$ resolution. Since we want to detect diver faces and subsequently the facial keypoints of the divers regardless of their physical distance from the robot, we employ a multi-scale learning approach as was used in~\cite{he2015spatial}. Therefore, we extract the initial features at three different scales, \ie from three different layers of ResNet-101. These features are further refined individually using a backbone network (we use either ResNet-50~\cite{he2016deep} or MobileNet-V2~\cite{sandler2018mobilenet} as the backbone). We refrain from using any pretrained weights for the backbone network to eliminate any learning biases from the ImageNet dataset. We then create a three-level ($P_1$, $P_2$, $P_3$) feature pyramid~\cite{lin2017feature} with these extracted features using both top-down and lateral connections, as described in~\cite{deng2020retinaface}. Instead of performing the final predictions directly on the feature pyramid, we use extra context learning modules to leverage information from the surrounding pixels. It is shown in~\cite{najibi2017ssh} that using such post-context learning blocks increases the receptive fields of the pixel locations, which enables knowledge from the surroundings and hence gives better inference. The outputs from the final three context modules have feature dimensions of $(80,80,256)$, $(40,40,256)$, and $(20,20,256)$, respectively. 

\subsection{Facial Anchors} \label{sec:anchor}
To detect the facial keypoints of the divers at different scales, we perform the network supervision at patch levels on the feature pyramid. These patches, called \textit{anchors}, were introduced in~\cite{wang2017face} for enabling multi-scale face detection. Since underwater HRI takes place at a reasonably close distance, we select $16$ as the smallest anchor size to detect the distant divers' faces. As for the rest, we use anchors of sizes $32, 64, 128,$ and $256$. Anchors $16$ and $32$ work on level $P_1$, $64$ and $128$ on level $P_2$, and finally the anchor $256$ works on level $P_3$. All the anchors have square dimensions. We consider an anchor to have a diver face if it matches with the ground truth bounding box with an Intersection over Union (IoU) value of at least $0.6$ and to have background if the IoU is less than $0.4$. All other anchors are ignored. 

\subsection{Objective Function Formulation}
One of the key objectives of this work is to accurately detect the facial keypoints, even when they are occluded from the camera’s perspective. Therefore, we formulate the objective function as a multi-loss optimization problem (as was used in~\cite{girshick2015fast}) where we not only want to detect diver faces and their facial keypoints but also maintain the underlying geometric relations among the $10$ facial keypoints. The individual loss components are described below:

\begin{enumerate}
    \item $\mathcal{L}_{fc}(f, f_{gt})$: This is a cross-entropy loss which governs whether a particular image patch (anchor) is a face or background. Here, $f$ is the probability of an anchor being a face. Additionally, if the ground truth anchor also contains a face, then the value of $f_{gt}$ is $1$, otherwise it is $0$. Here, $gt$ refers to the ground truth values.
    
    \item $\mathcal{L}_{fb}(\Vec{b}, \Vec{b}_{gt})$: This is the diver face bounding box regression loss where $\Vec{b}=[x_{min},y_{min},x_{max},y_{max}]$ is a vector containing the predicted face bounding box coordinates. Here, $\mathcal{L}_{fb}$ is a \textit{smooth-L$_1$ loss} which was introduced in~\cite{girshick2015fast} and is defined as, 
    \begin{equation}
        \text{smooth-L}_1(u) = \begin{cases}
                                    0.5u^2    \hspace{4ex}  \text{if $|u| \leq 1$} \\
                                    |u| - 0.5   \hspace{2ex}       \text{otherwise}
                               \end{cases} \label{eq:l1-loss}
    \end{equation}
    where $u$ is fed as the difference between the predicted and the ground truth bounding box coordinates. Finally, the loss is averaged for all four coordinates.
    
    \item $\mathcal{L}_{kp}(\Vec{p}, \Vec{p}_{gt})$: This is the diver face keypoints regression loss where $\Vec{p}=[x_1, y_1, \cdots, x_{10}, y_{10}]$. The loss is also calculated as a smooth-L$_1$ loss (as shown in Eq.~\ref{eq:l1-loss}). Essentially, $\mathcal{L}_{kp}$ supervises the algorithm to predict the $10$ facial keypoints closer to the ground truth values. For both $\mathcal{L}_{fb}$ and $\mathcal{L}_{kp}$ losses, the four bounding box coordinates and the $10$ facial keypoints values were normalized.
    
    \item $\mathcal{L}_{gm}(\Vec{p})$: Finally, we introduce a novel geometric loss term which enforces the algorithm to maintain the geometric relations among the keypoints. This should allow accurate keypoints regression even when all $10$ keypoints are occluded from the camera’s perspective. This is a vital criterion so that the network can accurately detect facial keypoints even when the diver is facing the robot at a $90$-degree angle. Since scuba masks are rigid objects, the points on a mask should never deviate from their relative positions (with respect to one another) in 3D space regardless of the head orientation of the diver. We leverage this idea on the image space and ensure that the outer six keypoints ($p_1, p_3, p_4, p_5, p_6, p_8$) on scuba mask are symmetric with respect to the middle two keypoints ($p_2, p_7$). We define the loss term as,
    \begin{align*}
        \mathcal{L}_{gm}(\Vec{p}) = \sum_{j = 2,7}{ l(p_{1j},p_{3j}) + l(p_{4j},p_{5j}) + l(p_{6j},p_{8j}) }
    \end{align*}
    where 
    \begin{equation*}
        l(p_{ab},p_{cb}) = |d_{p_{ab}}-d_{p_{cb}}|
    \end{equation*}
    \begin{equation*}
    \text{and, }    d_{p_{mn}} = \sqrt{(x_m-x_n)^2 + (y_m-y_n)^2}
    \end{equation*}
\end{enumerate}
In total, we minimize the following multi-loss objective function:
\begin{equation}
    \mathcal{L} = \mathcal{L}_{fc} + \alpha f_{gt} \mathcal{L}_{fb} + \beta f_{gt} \mathcal{L}_{kp} + \gamma f_{gt} \mathcal{L}_{gm} \label{eq:loss}
\end{equation}
where $\alpha, \beta, \gamma$ are set to $0.2$, $0.15$, and $0.1$, respectively. We use non-uniform loss weights to prevent overfitting~\cite{cortes2020agnostic}. 
A higher loss weight is assigned to $L_{fb}$ to ensure accurate diver face bounding box detection, which is crucial for robust keypoints regression. 
The last three loss terms in Eq.~\ref{eq:loss} are considered only if an anchor contains a face, \ie $f_{gt}=1$. Fig.~\ref{fig:model} shows the complete architecture of DATT-Net.

\section{Diver Attention Estimation} \label{spread_analysis}
We determine the attentiveness of a scuba diver using DATT-Net's prediction of the $10$ facial keypoints. First, we measure the spread of the outer keypoints on the scuba mask ($p_1, p_3, p_4, p_5, p_6, p_8$) with respect to the nose ($p_7$) of the diver. We only consider the $x$ coordinates of the points, as we found that including the $y$ coordinates does not improve performance of the estimation, except when a diver is looking completely upward or downward. However, the recognition module will always classify those postures as inattentive, which makes sense because a diver intending to interact will not be looking completely upward or downward. Therefore, we limit our computations in the $x$-axis. However, instead of directly comparing against the nose coordinate, we use the mean ($x_{avg}^{p_{2,7,9,10}}$) location of the inner keypoints ($p_2, p_7, p_9, p_{10}$) because these points should always lie on the same line in 3D space. To compute the spread in $x$ dimension, we use the following formulation,
\begin{equation*}
    x_{sprd} =  \frac{1}{6}\sum_{i=1,3,4,5,6,8}{\sqrt{ (x^{p_i} - x_{avg}^{p_{2,7,9,10}})^2 } }
\end{equation*}
This value is normalized with respect to the width of the predicted diver face bounding box. Second, we leverage the positions of the top three keypoints ($p_1, p_2, p_3$) on the diver's mask with respect to the center ($x_{c}$) of the face bounding box to define a pseudo-angle variable ($p_a$) to quantify the angular deviation of the diver's head from the center. We define,
\begin{equation*}
    p_a=x_{m}-x_{c}
\end{equation*}
where $x_{m}$ is the mean $x$ location of the keypoints $p_1, p_2, p_3$, and defines the center of the diver's mask. 

Finally, we determine the attentiveness ($att$) of a diver using the following conditions,  
\begin{equation*}
    att = \begin{cases}
                        True    \hspace{2.9ex}  \text{if $x_{sprd} > \lambda_1$ and $|p_a|<\lambda_2$} \\
                        False  \hspace{2ex}       \text{otherwise}
                    \end{cases}
\end{equation*}
where the values of $\lambda_1$ and $\lambda_2$ are empirically determined as $0.27$ and $10$, respectively, through testing on the $662$ images of the DATT test set. We also conclude that a diver is looking to their left if $p_a>0$ and right if $p_a<0$. The threshold criteria (\ie when to consider a diver as attentive or otherwise) involved in determining these values were set based on our extensive research with real robots and divers in numerous human-robot interaction tasks across diverse water bodies (pool, lake, and ocean). Our experiments show that these empirically determined values enable accurate diver attention estimation more than $92\%$ of the time.
\begin{figure}[t]
    \centering
    \vspace{2mm}
    \includegraphics[width=.99\linewidth]{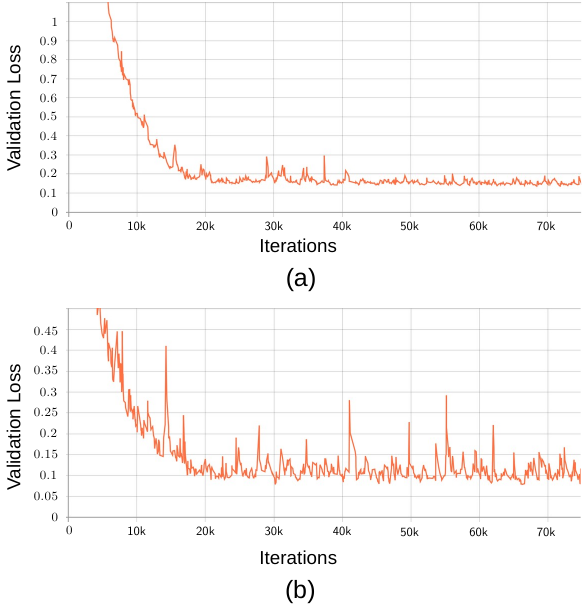}
    \caption{The training performance of DATT-Net in terms of the validation loss where the backbone is either (a) ResNet-50, or (b) MobileNet-V2. Here, iterations = $(\text{no. of training images}/\text{batch size})*\text{epochs}$.}
    \label{fig:loss}
    \vspace{-5mm}
\end{figure}

\section{Robot Controller}
We use the six-legged Aqua AUV platform~\cite{dudek2007aqua} to design a controller that uses the attentiveness of a diver to navigate and reorient itself for interaction. We create a Robot Operating System (ROS)~\cite{quigley2009ros} package consisting of two nodes. The first node continuously predicts the facial keypoints to compute the attentiveness score ($att$), while the second node keeps track of these values for the past $3$ seconds. Based on the recent attentiveness score (\textit{Attentive}, \textit{Looking Right}, or \textit{Looking left}), the controller intructs the Aqua AUV to try to catch the attention of the diver with a unique movement. If unsuccessful, the AUV autonomously navigates and reorients itself for interaction (\ie if the diver is looking right, the AUV does a left maneuver, and vice-versa). We implement the maneuvers using a motion controller introduced in~\cite{giguere2013wide-speed}. During the navigation-and-reorientation phase, the initial turn, the circular movement, and the final turn are governed by the pseudo-angle value, which is computed in the previous section. The ROS package and the implementation code for the controller are available at \url{https://github.com/IRVLab/datt_controller}.
\section{Experimental Evaluations}
To evaluate the performance of the diver attention estimation framework,
we prepare an evaluation set containing $812$ diver images. Of these, $662$ images are from the DATT test set and $150$ images are from multiple open-water environments. Furthermore, we deploy the proposed system on-board the Aqua AUV in real-world scenarios (robot trials carried out in a swimming pool and the Caribbean Sea) and evaluate the efficacy of the framework with scuba divers present in the field of view of the robot.   
To test the performance of the proposed robot controller, we check if the Aqua AUV successfully navigates and reorients itself with respect to the diver if they are found inattentive. 

\subsection{Implementation Details} \label{sec:implementation_details}
We use TensorFlow~\cite{abadi2016tensorflow} libraries to implement DATT-Net. The input images are resized to have $640\times 640$ resolutions. To augment the DATT dataset, we use the following schemes (as suggested in~\cite{howard2013some, zhang2017single}): random cropping, image flipping, normalization, image distortion (varying the brightness, hue, saturation, contrast, and sharpness), etc. During training, we use a batch size of $8$, learning rate of $10^{-2}$ with a scheduler that lowers the value down to $10^{-5}$, and Stochastic Gradient Descent (SGD) as the optimizer with a momentum of $0.9$. We train the network using $2,652$ training images on an Nvidia GeForce RTX 2080 GPU for $300$ epochs where we notice convergence in the validation loss for both ResNet-50 and MobileNet-V2 backbones (see Fig.~\ref{fig:loss}). 
For on-the-bench evaluations, we used an Intel\textregistered{} Xeon\textregistered{} E5-2650 CPU to run DATT-Net. During real-world deployments in both closed- and open-water environments, we used the on-board computing platform (Nvidia Jetson\textregistered{} TX2) of the Aqua AUV.
\begin{table}[t]
\centering
\vspace{2mm}
\caption{Diver Face detection and Keypoints regression results on the evaluation set. The values are in percentages.}
\begin{tabular}{lccc}
  \toprule
  \multirow{3}{*}{\textbf{\begin{tabular}[c]{@{}c@{}}\textbf{Method}\end{tabular}}}   & \multicolumn{2}{c}{\textbf{Face Det.}} & \textbf{Keypoints Reg.}\\
  & \textbf{AP} & \textbf{mAP} & \textbf{PCK}\\
  & ($\uparrow$) & ($\uparrow$) & ($\uparrow$)\\
  \midrule
  PyramidBox~\cite{tang2018pyramidbox} &  $61.45$        &   $44.83$     &     -              \\
  BlazeFace~\cite{bazarevsky2019blazeface} &  $66.95$        &   $45.86$      &     -             \\
  DSFD~\cite{li2019dsfd} &  $77.22$        &   $54.70$           &     -        \\
  RetinaFace~\cite{deng2020retinaface} &  $70.81$ &  $48.33$     & -               \\ 
  \textbf{DATT-Net (+ ResNet-50)} &  $\mathbf{91.75}$ &  $\mathbf{72.56}$ &  $\mathbf{86.21}$ \\
  Same w/o $\mathcal{L}_{gm}$ & $82.64$ & $55.22$ & $71.79$\\ 
  DATT-Net (+ MobileNet-V2) &  $87.93$ &      $55.33$ & $79.88$ \\
  Same w/o $\mathcal{L}_{gm}$ &  $75.37$ &      $50.22$ &  $64.44$ \\ 
  \bottomrule
\end{tabular}
\label{tab:face_detect_kp}
\vspace{-4mm}
\end{table}

\subsection{Results}
\subsubsection{Diver Face Detection and Keypoints Regression}
Since a major component of the proposed framework is diver face detection, we start by comparing the diver face detection performance of DATT-Net against the State-Of-The-Art (SOTA) face detection algorithms.
\textit{Metrics:} we compute the Average Precision (AP) for IoU=$0.5$ and Mean Average Precision (mAP) for IoU=[$0.5$:$0.05$:$0.95$].
The results are presented in Table~\ref{tab:face_detect_kp}. From the table, it is evident that DATT-Net achieves significantly better performance than the SOTA methods, regardless of the backbone network. 
While we also considered LFFD~\cite{he2019lffd}, errors present in the available GitHub code prevented us from training the model for quantitative comparisons. 
Even though RetinaFace~\cite{deng2020retinaface} and BlazeFace~\cite{bazarevsky2019blazeface} use facial landmarks to robustly detect faces, their accuracy suffer on diver faces. This might be due to divers wearing different types of scuba gear because these methods worked well for divers who did not wear masks, snorkels, or regulators (\eg wearing only pairs of goggles). Furthermore, it is also evident from the table that the use of the additional supervision in the form of geometric loss contributes to better diver face detection because the AP value decreases when the geometric loss is excluded during training. DATT-Net achieves the best performance (AP=$91.75\%$, mAP=$72.56\%$) when ResNet-50 is used as the backbone network. 
\begin{figure*}[t]
    \centering
    \vspace{2mm}
    \includegraphics[width=.95\linewidth]{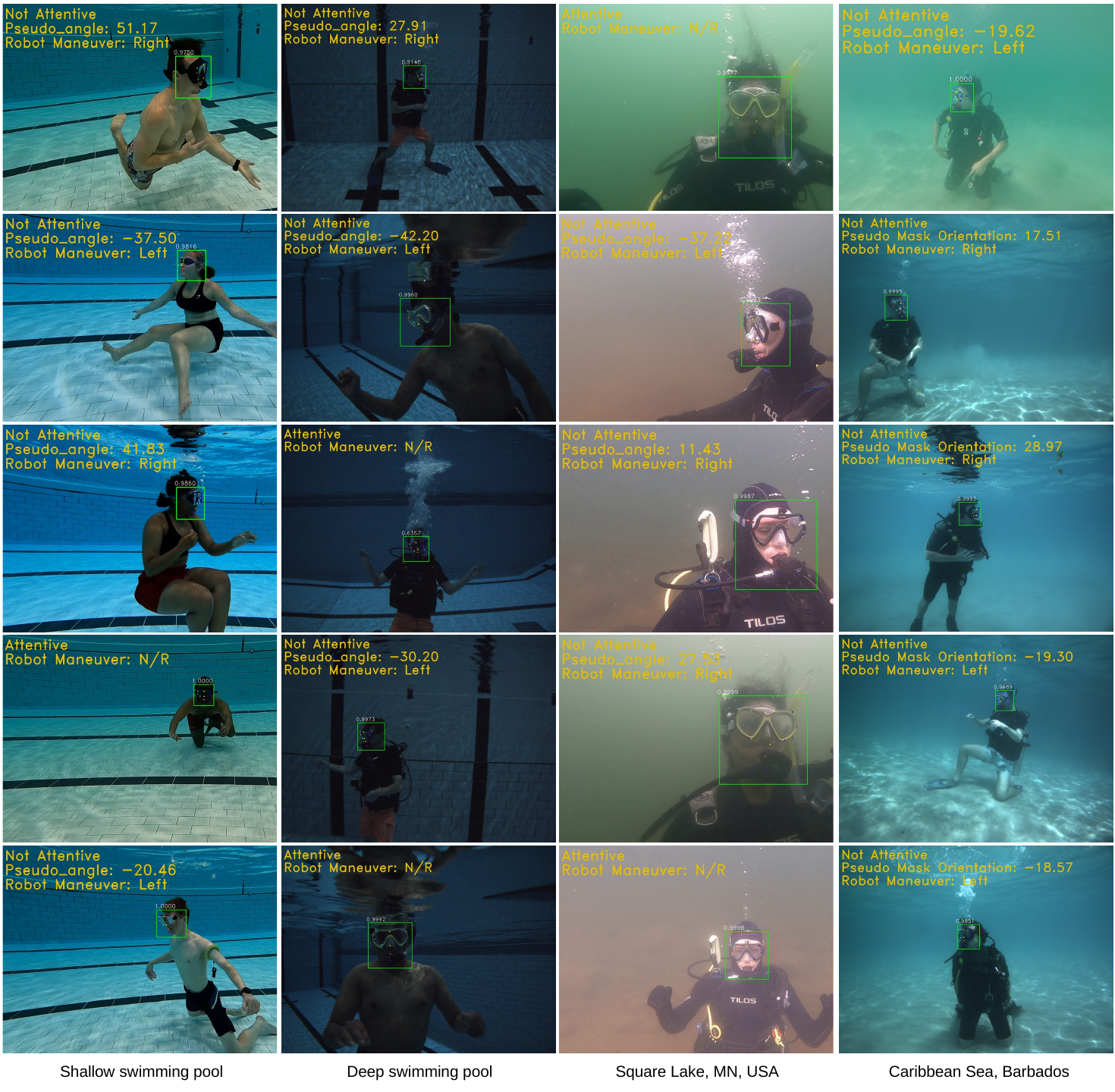}
    \caption{Qualitative performance of the proposed diver attention estimation framework when run on-board the Aqua AUV in both closed- and open-water environments. Note the robustness of our method as it works at different distances and in challenging lighting conditions.}
    \label{fig:qual}
\end{figure*}

For the keypoints regression task, we are only able to evaluate the performance of DATT-Net models because we do not have ground truth values for BlazeFace and RetinaFace, whereas PyramidBox~\cite{tang2018pyramidbox} and DSFD~\cite{li2019dsfd} do not output facial keypoints.
\textit{Metric:} we compute the Percentage of Correct Keypoints (PCK)~\cite{pishchulin2016deepcut} by measuring if the predicted keypoint and the ground truth are within a certain distance threshold. For our purpose, we set the threshold as $10\%$ of the mask width (distance between the keypoints \#4 and \#5). From Table~\ref{tab:face_detect_kp}, we see that DATT-Net with ResNet-50 backbone achieves the best PCK value of $86.21\%$. As before, we notice similar phenomenon where the PCK value decreases when the geometric loss is excluded.

\subsubsection{Diver Attention Estimation}
First, we perform qualitative evaluations on the performance of the proposed diver attention estimation technique. We test our algorithm on the evaluation set and also in two open-water trials (Square Lake, MN, USA, and Caribbean Sea, off the coast of Barbados). Fig.~\ref{fig:qual} presents the qualitative results. From the figure we see that the proposed algorithm accurately determines the attentiveness of divers irrespective of the type of scuba gear they are wearing. It is also evident that the proposed solution works at different distances and in various lighting conditions. This is understandable because we have used multi-scale learning (using different sizes of anchors) on a feature pyramid and image distortion as a data augmentation scheme, respectively, during the training of DATT-Net (as described in Sec.~\ref{sec:anchor} and Sec.~\ref{sec:implementation_details}). It is important to note that even though DATT-Net was trained using images collected from closed-water environments, \eg  swimming pools, it performs well on both lake and sea images captured by the robot's camera. The framework also performs well even when the majority of the diver's face is not visible (\eg because they are facing the AUV at a $90$-degree angle). This points to DATT-Net indeed benefiting from learning the geometry of the keypoints.

Second, we quantitatively evaluate the performance of the proposed framework. To our knowledge, there is no other research tackling the problem of diver attention estimation that we can compare against. Consequently, we decide to use terrestrial facial keypoints regression techniques as our baselines, such as a CNN-based keypoints regressor~\cite{newell2016stacked}, BlazeFace, and the Perspective-n-Point (PnP) algorithm~\cite{fischler1981random}. To determine the attentiveness of divers using these techniques, we do the following. 
First, we compute the facial keypoints from diver images in our evaluation set using the CNN-based keypoints regressor, BlazeFace, and DATT-Net. Then, we use spread analysis (as discussed in Sec.~\ref{spread_analysis}) to determine the attentiveness of divers. We had to modify which keypoints to consider during the calculation of the spread analysis for the first two techniques as the keypoints' locations are defined differently for them.
In contrast, we determine the diver attention for the PnP technique as follows. We add depth information to the facial keypoints to prepare a 3D model of the diver's head wearing a mask and snorkel. 
The depth values are assigned relative to the mask, with points vertically above the mask having a positive depth, and those vertically below having a negative value, with the depth at mask being zero.
Now, given a diver image, we run DATT-Net to find the $10$ facial keypoints and use the 3D model of the diver's head to estimate the head pose of the diver (\ie translation and rotation vectors) using the PnP technique. From the translation and rotation vectors of the nose (keypoint \#7), we compute its angular deviation from the center and determine the attentiveness of the diver.
Table~\ref{tab:attention_est} presents the results. We see that our framework achieves a superior accuracy of $89.41\%$ compared to the other baselines.

\subsubsection{Robot Controller's Performance}
We deployed our robot controller on-board the Aqua AUV using ROS and tested the performance of the controller in both closed- and open-water environments. The experimental setup involved two divers and one AUV, with one diver working as a robot wrangler while the other served as the interaction target. When the AUV detected the diver to be inattentive, it initiated a unique movement pattern to draw their attention. If unsuccessful, the AUV autonomously devised an optimal trajectory to navigate and reposition itself for effective interaction with the diver. This planning process relied on the $pseudo\_angle$ value derived from diver attention estimation. In contrast, if the diver is found to be attentive, the AUV acknowledged with another unique movement pattern and began the interaction process. The performance of the robot controller in real-world scenarios is provided in the accompanying video, showcasing its effectiveness in facilitating AUV-diver interaction.
\begin{table}[t]
\centering
\vspace{2mm}
\caption{Diver attention estimation results on the evaluation set. The accuracy values are in percentages.}
\begin{tabular}{l|cccc}
  \toprule
  \textbf{Method}  & CNN~\cite{newell2016stacked} & BlazeFace~\cite{bazarevsky2019blazeface} & PnP~\cite{fischler1981random} & \textbf{Ours} \\
   \textbf{Accuracy} ($\uparrow$) & $38.30$ & $42.73$ &  $45.07$    &     $\mathbf{89.41}$        \\ \bottomrule 
\end{tabular}
\label{tab:attention_est}
\vspace{-4mm}
\end{table}

\subsection{Limitations of the Proposed Framework}
\begin{figure}[t]
    \centering
    \vspace{2mm}
    \includegraphics[width=.96\linewidth]{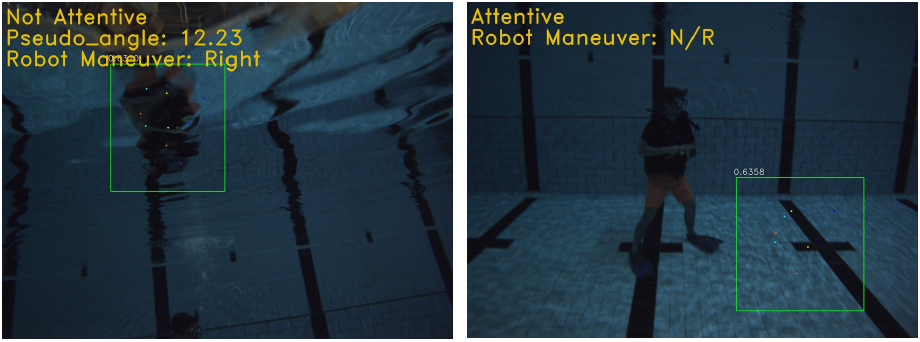}
    \caption{Two examples of diver attention estimation failures where the algorithm incorrectly detects a diver face in the reflection on the water surface (left image) and in the floor of the pool (right image).}
    \label{fig:limitation}
    \vspace{-4mm}
\end{figure}

Our method relies on the prediction of the diver face bounding box. If the predicted bounding box does not include any faces, then all the subsequent modules fail. Although the performance of DATT-Net in detecting diver faces is quite robust, we still see a few failure cases. As seen in Fig.~\ref{fig:limitation}, the algorithm mistakenly found a diver face in the water reflection and also mistook a lane marking as a face on the swimming pool floor. Also, the current robot controller is designed to only change its yaw value during the navigation-and-reorientation phase. That is, our controller makes an assumption that the interaction will happen at a fixed height in the water column.
\section{Conclusions}
In this work, we have presented the design and implementation of a diver attention estimation framework to facilitate effective underwater HRI. With on-the-bench experiments, closed- and open-water robot trials, we have shown that our proposed system determines the attentiveness of divers with promising accuracy. Our robot controller uses the attentiveness score to allow the AUV to navigate and reorient itself before initiating interaction. The key advantages of our method are that it works on physical AUVs, detects the attentiveness of divers using only monocular cameras, works at various distances and lighting conditions, does not require knowledge of the global positions of divers to function, and is invariant to different types of scuba gear. This framework will serve as a step toward learning the overall intent of a scuba diver, allowing an AUV to make more informed decisions when collaborating with a diver. As immediate next steps, we will tune our robot controller so it can plan the navigation-and-reorientation task more efficiently, with full 6-degrees-of-freedom control of the robot's motion. We also hope to account for divers' activities as a consideration before the robot initiates interaction.


\bibliographystyle{ACM-Reference-Format}
\bibliography{root}


\end{document}